\documentclass[final]{colt2025}

\title[Translation-Equivariance of Normalization Layers]{Translation-Equivariance of Normalization Layers and Aliasing \\ in Convolutional Neural Networks}
\usepackage{times}

\coltauthor{%
 \Name{J\'er\'emy Scanvic} \Email{jeremy.scanvic@ens-lyon.fr}\\
 \addr Laboratoire de Physique, ENS de Lyon (LPENSL)\\
 \addr Foxstream, Vaulx-en-Velin, France%
 \AND
 \Name{Quentin Barth\'elemy} \Email{q.barthelemy@foxstream.fr}\\
 \addr Foxstream, Vaulx-en-Velin, France%
 \AND
 \Name{Juli\'an Tachella} \Email{julian.tachella@ens-lyon.fr}\\
 \addr Laboratoire de Physique, ENS de Lyon (LPENSL)%
}

\usepackage{array}
\usepackage{textcomp}
\usepackage{stfloats}
\usepackage{verbatim}
\usepackage{standalone}
\usepackage{hyperref}
\usepackage{amsfonts}
\usepackage{booktabs}
\usepackage{chngpage}
\usepackage[capitalize]{cleveref}
\usepackage{orcidlink}

\usepackage{thmtools, thm-restate}

\allowdisplaybreaks

\begin{document}

\maketitle

\begin{abstract}
	The design of convolutional neural architectures that are exactly equivariant to continuous translations is an active field of research. It promises to benefit scientific computing, notably by making existing imaging systems more physically accurate. Most efforts focus on the design of downsampling/pooling layers, upsampling layers and activation functions, but little attention is dedicated to normalization layers. In this work, we present a novel theoretical framework for understanding the equivariance of normalization layers to discrete shifts and continuous translations. We also determine necessary and sufficient conditions for normalization layers to be equivariant in terms of the dimensions they operate on. Using real feature maps from ResNet-18 and ImageNet, we test those theoretical results empirically and find that they are consistent with our predictions\footnote{The code for our experiments is available at \url{https://github.com/jscanvic/normalization-layers}}.
\end{abstract}

\section{Introduction}

Convolutional neural networks have long been thought to be equivariant to translations thanks to the use of convolutional layers. It is now understood that regular layers used in most convolutional networks are prone to aliasing and that this aliasing breaks the equivariance to translations~\citep{azulay19Why,zhang19Making,zou20Delving}. This is especially the case of downsampling/pooling layers, upsampling layers and activation functions.

The first attempts at fixing the problem came in the form of layers featuring anti-aliasing filters, this is the case of blur pooling~\citep{zhang19Making,zou20Delving,michaeli23AliasFree}, of filtered activation functions, including filtered ReLU~\citep{karras21AliasFree} which is less prone to aliasing than the traditional ReLU, and filtered polynomial activation functions~\citep{michaeli23AliasFree} which are perfectly free of aliasing. Other works have focused on the design of networks equivariant to discrete translations (known as shifts) using adaptive downsampling and upsampling layers~\citep{chaman21Truly,chaman21TrulyEquivariant,kim23ModelAgnostic}, possibly with learnable parameters~\citep{rojas-gomez22Learnable,saha24Improving}. While these approaches guarantee perfect equivariance to shifts, they do not cover full equivariance to continuous translations.

Contrary to the other layers, little has been said on the translation-equivariance of normalization layers. Most works use standard batch normalization layers that are by far the most popular normalization layers in convolutional neural networks \citep{chaman21Truly,chaman21TrulyEquivariant}, but little is said about their equivariance. In their recent work, \citet{michaeli23AliasFree} adapt the modern ConvNext architecture \citep{liu22ConvNet} to make it equivariant. In particular, they claim that the normalization layers used in the original architecture are not equivariant to translations, and they propose an equivariant alternative.

In this work, we shed light on what makes certain normalization layers equivariant to shifts and translations. Using a new theoretical framework that covers the most common normalization layers, we show that dividing by the standard deviation and applying an affine transform are the two steps that might cause a loss of equivariance. On the other hand, subtracting the mean poses no problem. We validate our theoretical results empirically using real feature maps obtained from a network pre-trained on ImageNet~\citep{deng09ImageNet}.

\begin{table}[htbp]
	\centering
	\setlength{\tabcolsep}{8pt} % default: 6pt
	\begin{tabular}{l|c|c|c|c}
	\toprule
	Layer & Centering & Scaling & Affine & Equivariance \\
	\midrule
	BatchNorm~\citep{ioffe15Batch} & $B$, $H$, $W$ & $B$, $H$, $W$ & $C$ & Translation \\
	InstanceNorm~\citep{ulyanov17Instance} & $H$, $W$ & $H$, $W$ & None & Translation \\
	LayerNorm-CHW~\citep{ba16Layer} & $C$, $H$, $W$ & $C$, $H$, $W$ & $C$, $H$, $W$ & Neither \\
	LayerNorm-C~\citep{liu22ConvNet} & $C$ & $C$ & $C$ & Shift \\
	LayerNorm-AF~\citep{michaeli23AliasFree} & $C$ & $C$, $H$, $W$ & $C$ & Translation \\
	\bottomrule
	\end{tabular}
	\caption{\textbf{Equivariance of normalization layers.} Normalization layers consist in three steps: a centering step, a scaling step, and a learned affine step. Depending on the layer, the steps are performed on different dimensions (batch $B$, channels $C$, height $H$ and width $W$). We show theoretically and empirically in~\Cref{section:analysis,section.experiments} that equivariance to discrete shifts requires the affine step not to operate on the spatial dimensions $H, W$, and for equivariance to continuous translations, that the scaling step operates at least on the spatial dimensions.}
	\label{tbl:normalization_layers}
\end{table}

Our contributions are the following:
\begin{itemize}
	\item We propose a new theoretical framework for understanding the equivariance of normalization layers to shifts and translations.
	\item We present necessary and sufficient conditions for a normalization layer to be equivariant to discrete shifts, and to continuous translations.
	\item We validate our theoretical results by measuring and comparing the equivariance of five normalization layers using real feature maps.
\end{itemize}

\section{Related work}

\paragraph{Alias-free layer norm} \citet{michaeli23AliasFree} adapt the ConvNext architecture~\citep{liu22ConvNet} to make it equivariant to continuous translations, and propose an alternative translation-equivariant normalization layer. Indeed, they claim that the original normalization layer, channel-wise layer normalization, is not equivariant to translations due to aliasing in the scaling step. In order to alleviate this problem, they change the dimensions the standard deviation is computed on from just the channel dimension to the channel and spatial dimensions. In this work, we prove that their claim is correct and that their solution is valid by showing that their proposed layer is indeed equivariant to translations, while the original one is only equivariant to shifts.

\paragraph{Steerable layers} Many works focus on adapting convolutional layers to larger classes of equivariance~\citep{cohen16Group,cohen16Steerable}. Indeed, while convolutional layers are equivariant to discrete shifts and even to continuous translations, they are not equivariant to other transformations like rotations and flips. Using parameter-sharing schemes~\citep{ravanbakhsh17Equivariance}, they attain perfect equivariance to discrete transformations, e.g., 90\textdegree{} rotations, but they do not generally attain perfect equivariance to continuous transformations, e.g., to continuous translations and rotations at once, due to fundamental limitations related to aliasing and sampling theory~\citep{weiler21General}. In this work, we focus on the (non-linear) normalization layers and on their equivariance to shifts and translations, and whether they are equivariant to other transformations goes beyond this scope.

\section{Background}
\label{section:background}

Normalization layers in convolutional neural networks take in feature maps $x \in \mathbb R^{N}$ and compute a normalized feature map $f_\theta(x) \in \mathbb R^{N}$. Here $N = B \times C \times H \times W$, where $B$ denotes the batch size, $C$ the number of channels, and $H$ and $W$ the height and width of the feature map. The variable $\theta \in \mathbb R^p$ denotes the learnable parameters of the layer. In this section, we introduce the mathematical background behind equivariance to shifts and translations, and the relation between equivariance to continuous translations and aliasing.

\paragraph{Shifts and translations} In many applications, circular translations are used to move the content of feature maps around without loosing information at the boundary~\citep{zhang19Making,michaeli23AliasFree}. Usually, the displacement is assumed to span a whole number of pixels in both directions, e.g., 2 px down and 3 px to the right, and in that case, the (discrete) translation is referred to as a shift. The discrete shift operator $T_g$ can be understood as a simple permutation of the pixels and it is defined as
\begin{equation}
	(\mathop{T_g} x)_{bhcw} = x_{b,c,(h - h')_H, (w - w')_W},
\end{equation}
where $g = (h', w') \in \mathbb Z^2$ is the displacement vector. The notations $( \cdot )_H$, and $( \cdot )_W$ denote the remainder of an integer modulo $H$ and $W$, respectively, and they are what makes the shift operator circular. The indices $b = 0, \ldots, B - 1$, $c = 0, \ldots, C - 1$, $h = 0, \ldots, H - 1$ and $w = 0, \ldots, W - 1$ each correspond to one of the four dimensions in a batch of feature maps.

Discrete shifts are sometimes insufficient and finer sub-pixel translations need to be considered, notably when studying texture sticking in certain generative models~\citep{karras21AliasFree}. In that case, discrete feature maps are generally assumed to be the sampling of a latent continuous feature map, similarly to how discrete pictures are sampled from incoming continuous images hitting the camera sensor. Even though there is generally a loss of information during sampling, Shannon's sampling theorem~\citep{vetterli14Foundations} guarantees that the low-frequency information is preserved as long as proper anti-aliasing is done during sampling, and that the corresponding bandlimited continuous signal can be recovered using sinc interpolation.

Guided by this underlying assumption, continuous translation is generally defined as the succession of three steps: i) interpolation into a continuous image, ii) continuous translation of the continuous image, and iii) sampling of the translated image back to the original grid~\citep{karras21AliasFree,michaeli23AliasFree}. For sinc interpolation, it amounts to first applying the discrete Fourier transform ($\mathrm{DFT}_2$), the right phase shift, and then the inverse discrete Fourier transform ($\mathrm{IDFT}_2$)~\citep{vetterli14Foundations}
\begin{equation}
	(\mathop{T_g} x)_{bchw} = \mathrm{IDFT}_2\left(e^{-i 2 \pi \left( \frac{hh'}{H} + \frac{ww'}{W} \right)} \mathrm{DFT}_2(x_{bchw})\right),
\end{equation}
where $g = (h', w') \in \mathbb R^2$ is a displacement vector that might not span a whole number of pixels in both directions.

Even though it might not be obvious from the formula, continuous translations coincide with discrete shifts for whole pixel displacements, a fact known as the shift theorem~\citep{vetterli14Foundations}.

\paragraph{Equivariance} Equivariant functions are functions whose output is translated accordingly with its input when it is translated. The layer $f_\theta : \mathbb R^N \to \mathbb R^N$ is equivariant to shifts $\mathcal G = \mathbb Z^2$, or translations $\mathcal G = \mathbb R^2$, if it satisfies
\begin{equation} \label{eq:equivariance}
	f_\theta(T_g x) = T_g f_\theta(x),\ \forall \theta \in \mathbb R^p, \forall g \in \mathcal G,\ \forall x \in \mathbb R^N,
\end{equation}
We emphasize that the equivariance needs to be satisfied not only for all inputs $x$ and displacements $g$, but also for all sets of parameters $\theta$. We also refer to this property as architectural equivariance, as opposed to learned equivariance~\citep{gruver24Lie}.

\paragraph{Aliasing} Equivariance to translations can only be satisfied if equivariance to shifts is satisfied in the first place. This is because discrete shifts are a special case of continuous translations. Of course, this is generally not a sufficient condition and aliasing is key to determine when shift-equivariant functions are equivariant to translations.

The same way discrete feature maps can be understood as continuous feature maps through sinc interpolation, functions operating on discrete feature maps can also be understood as functions operating on continuous feature maps. The idea is that given a function operating on discrete feature maps, it is possible map an input continuous feature map to an output feature map by: i) first sampling the input to obtain a discrete feature map, ii) then applying the discrete function, and iii) to interpolate the resulting discrete feature map back to get the output continuous feature map.

\citet{karras21AliasFree} give multiple examples of discrete shift-equivariant functions that are associated to continuous translation-equivariant functions. For instance, convolutions and point-wise activation functions like ReLU. They also show that the only possible cause of a lack of translation-equivariance for the discrete function is if the continuous function introduces energy above the Nyquist frequency. In that case, the energy folds back into the lower frequencies in the discrete case (aliasing), causing a loss of translation-equivariance. This is notably what happens for ReLU, which is not equivariant to continuous translations.

\section{Analysis of normalization layers}
\label{section:analysis}

Normalization layers have been introduced to improve the training dynamics of deep neural networks~\citep{ioffe15Batch}. They generally consist in three steps:
\begin{enumerate}
	\item A centering step
		\begin{equation}
			x \mapsto x - \mathbb E[x]
		\end{equation}
	\item A scaling step
		\begin{equation}
			x \mapsto \frac x {\sqrt{\mathop{\mathrm{Var}}(x)}}
		\end{equation}
	\item A learned affine transform with parameters $\theta = [\gamma; \beta]$
		\begin{equation}
			x \mapsto \gamma \odot x + \beta
		\end{equation}
\end{enumerate}
where $\odot$ is the Hadamard product. We do the analysis for feature maps with a non-vanishing variance, and we leave out the tiny $\varepsilon$ that is generally added to the variance in practice to avoid divisions by zero.

The differences in the different normalization layers lie in the dimensions ($B$, $C$, $H$ and/or $W$) on which the statistics are computed for centering and scaling, and in the dimensions on which the affine transform operates, if an affine transform is present.

In this work, we focus on five normalization layers, four standard ones and one designed to be equivariant: batch normalization~(BatchNorm)~\citep{ioffe15Batch}, instance normalization~(InstanceNorm)~\citep{ulyanov17Instance}, layer normalization on the whole feature map~(LayerNorm-CHW)~\citep{ba16Layer,wu18Group}, layer normalization on the channels~(LayerNorm-C)~\citep{liu22ConvNet}, and alias-free layer normalization~(LayerNorm-AF)~\citep{michaeli23AliasFree}. Their respective definitions are summarized in~\Cref{tbl:normalization_layers} and are presented in more details below.

Except for LayerNorm-AF, all other normalization layers perform centering and scaling on the same dimensions. For BatchNorm, they are the batch and spatial dimensions $(B,H,W)$, for InstanceNorm, they are the spatial dimensions $(H, W)$, for LayerNorm-C, it is the channel dimension $(C)$, and for LayerNorm-CHW, it is the channel and spatial dimensions $(C,H,W)$. On the other hand, LayerNorm-AF performs its centering step on the channel dimension $(C)$, and its scaling step on the channel and spatial dimensions $(C,H,W)$. In terms of affine step, InstanceNorm has none, BatchNorm, LayerNorm-C and LayerNorm-AF have one that operates on the channel dimension $(C)$, and LayerNorm-CHW has one that operates on the channel and spatial dimensions $(C,H,W)$.

In order to understand what causes a loss of equivariance in a function comprised of multiple independent steps, it is customary to study the equivariance of each step separately. Indeed, the composition of multiple functions is equivariant as long as each function is itself equivariant~\citep{michaeli23AliasFree}. We apply this reasoning here to determine necessary and sufficient conditions for the equivariance of normalization layers to shifts and translations.

Discrete shifts act as pixel permutations as mentioned in~\Cref{section:background},
and shifting input feature maps results in statistics shifted accordingly. The shifted feature maps are subtracted and divided entry-wise by the shifted statistics, resulting in an overall shifted output before the affine step. At this point, either there is no affine step and the whole normalization layer is equivariant to shifts, or there is one and its equivariance is the equivariance of the layer. Unlike in the scaling step, which also consists in an entry-wise multiplication/division, only one of the two factors is ever shifted. Indeed, the standardized feature map shifts along with shifts in the input, but the learned affine matrix is fixed. In this regard, the affine step is equivariant if, and only if, it scales every pixel similarly no matter its position in the image plane. Said otherwise, if, and only if, it does not operate in the spatial dimensions. \Cref{theorem:shifts} follows directly and is proven in more details in~\Cref{section:proofs}.

\begin{restatable}{theorem}{theoremShifts} \label{theorem:shifts}
	A normalization layer is equivariant to discrete shifts if, and only if, its affine step does not operate on the spatial dimensions, or if has no affine step altogether.
\end{restatable}

In general, it can be difficult to determine if a discrete function is equivariant to continuous translations or not. Fortunately, there are cases for which it is significantly easier, and the present case is one of them. As explained in~\Cref{section:background}, equivariance to translations is only satisfied if equivariance to shifts is satisfied as well. As a result, it follows from~\Cref{theorem:shifts} that a normalization layer is equivariant to translations only if its affine step does not operate on the spatial dimensions, or if it has no affine step altogether. We assume that this condition is verified in the remainder of this section. Moreover, discrete functions that are equivariant to shifts, and that are also associated to a translation-equivariant continuous function, are themselves equivariant to translations if, and only if, they do not cause aliasing, or an increase in bandwidth past the Nyquist frequency in the continuous domain. Using this characterization, we determine a necessary and sufficient condition for normalization layers to be equivariant to translations.

The arguments that we use to prove, under the right condition, the shift-equivariance of the centering, scaling and affine steps in normalization layers in the discrete seeting also prove that they are translation-equivariant in the continuous setting. Continuous translations operate as a permutation of pixels in continuous images, the same way discrete shifts operate as a permutation of pixels in discrete images. Normalization layers are thus equivariant to translations in the continuous setting, and they are also equivariant to translations in the discrete setting if, and only if, it does not increase the bandwidth of bandlimited feature maps in the continuous case.

Normalization layers consist entirely in entry-wise additions and multiplications, which transform, in the Fourier domain, into entry-wise additions and spatial convolutions~\citep{vetterli14Foundations}. Entry-wise addition does not increase the bandwidth of bandlimited feature maps, but convolution with spatially varying kernels does: the resulting bandwidth being the sum of the bandwidths of the two convoluted images~\citep{vetterli14Foundations}. In the case of normalization layers, the only entry-wise multiplication of the input signal by a spatially varying kernel is in the scaling step, and only if the standard deviation is not computed at least on the spatial dimensions. \Cref{theorem:translations} follows directly and is proven in more details in~\Cref{section:proofs}.

\begin{restatable}{theorem}{theoremTranslations} \label{theorem:translations}
	A normalization layer is equivariant to continuous translations if, and only if, it is equivariant to shifts, and the standard deviation is computed at least on the spatial dimensions.
\end{restatable}

A direct corollary of~\Cref{theorem:shifts,theorem:translations} is that the dimensions on which the centering statistics are computed are irrelevant to the overall equivariance of the normalization layer to shifts and translations.

Our new theoretical results highlight the importance of the choice of normalization layer when designing equivariant neural architectures. In particular, they predict that the layer normalization used in the ConvNext architecture~\citep{liu22ConvNet} is equivariant to shifts, but not to translations, hindering the equivariance of the whole architecture. Batch normalization, on the other hand, is perfectly equivariant to both shifts and translations. In~\Cref{section.experiments}, we further show that those theoretical predictions hold empirically as well.

\section{Experiments}
\label{section.experiments}

In order to measure and compare the equivariance of the five different normalization layers mentioned in~\Cref{tbl:normalization_layers}, we define and compute the average equivariance error of each layer. We measure the equivariance of the normalization layers as the error corresponding to~\cref{eq:equivariance}:
\begin{equation}
	e = \mathop{\mathbb E}_{x, \gamma, \beta, g} \Big[ d(f_{\gamma,\beta}(\mathop{T_g} x), \mathop{T_g} f_{\gamma,\beta}(x)) \Big]
\end{equation}
where $d(\cdot, \cdot)$ is the cosine distance, defined in~\cref{eq:cosine_distance}. We compute the equivariance error for two different transform distributions, resulting in two distinct equivariance errors: $e_{\mathrm T}$ for translations, and $e_{\mathrm S}$ for shifts. For translations, the displacement parameter $g$ is sampled uniformly from $[0, H) \times [0, W)$, and for shifts it is uniformly sampled and from $\{ 0, \ldots, H - 1 \} \times \{ 0, \ldots, W - 1 \}$. The cosine distance is a standard metric for comparing two feature maps~\citep{zhang19Making}, and it is defined as:
 \begin{equation} \label{eq:cosine_distance}
	d(x, y) = 1 - \frac 1 {BHW} \sum_{b = 0}^{B - 1} \sum_{h = 0}^{H - 1} \sum_{w=0}^{W - 1} \frac{\sum_{c=0}^{C - 1} x_{bchw} y_{bchw}}{\sum_{c=0}^{C - 1} |x_{bchw}|^2  \sum_{c=0}^{C - 1}|y_{bchw}|^2}.
\end{equation}

For the learnable parameters, we use the combination of two distributions i) default initialization, i.e., $\gamma = \mathbf{1}$ and $\beta = \mathbf{0}$, and ii) Gaussian initialization with mean $0$ and standard deviation $1$ to simulate learned parameters. BatchNorm behaves differently in training mode and eval mode so we randomize its mode as well, and we also do a separate experiment with fixed mode to see if there is a significant difference. The results of this separate experiment are listed in~\Cref{tbl:batch_norm}. The other norms behave the same in both modes and we leave them in evaluation mode.

The feature maps are sampled from real feature maps obtained using the 50,000 validation images of ImageNet~\citep{deng09ImageNet} and a pre-trained ResNet-18~\citep{he15Deep}. They are obtained by feeding in batches of 1024 images to the network, and gathering the feature maps passed as input to each of the 20 batch normalization layers in the network, resulting in about 1,000 (batched) feature maps.

\begin{table}[bthp]
	\centering
	\setlength{\tabcolsep}{28pt} % default: 6pt
	\begin{tabular}{lll}
	\toprule
	 & Shifts & Translations \\
	Layer &  &  \\
	\midrule
	BatchNorm & \textbf{9.23e-09 ± 2.86e-12} & \textbf{1.28e-06 ± 1.70e-09} \\
	InstanceNorm & \textbf{9.58e-09 ± 5.29e-12} & \textbf{7.13e-06 ± 1.86e-08} \\
	LayerNorm-CHW & 4.97e-01 ± 3.54e-04 & 4.97e-01 ± 3.53e-04 \\
	LayerNorm-C & \textbf{4.66e-09 ± 7.05e-12} & 2.44e-03 ± 2.66e-06 \\
	LayerNorm-AF & \textbf{9.17e-09 ± 4.08e-12} & \textbf{8.08e-07 ± 2.59e-09} \\
	\bottomrule
	\end{tabular}
	\caption{\textbf{Equivariance error of normalization layers.} The equivariance error of each layer is computed using feature maps obtained from ResNet-18 and ImageNet. The metric is the cosine distance between feature maps transformed before, and after the normalization layer. It is lower for more equivariant layers. The results are consistent with the theoretical predictions shown in~\Cref{tbl:normalization_layers}. In \textbf{bold}, values lower than $10^{-4}$. Values: avg $\pm$ s.e.}
	\label{tbl:main}
	\vspace{-1cm}
\end{table}

In \Cref{tbl:main}, there are two clusters of layers for equivariance to shifts: those with an error in the range from $10^{-10}$ to $10^{-9}$, namely BatchNorm, InstanceNorm, LayerNorm-C and LayerNorm-AF; and the remaining one higher in the range from $10^{-2}$ to $10^{-1}$, namely LayerNorm-CHW. In terms of equivariance to translations, there are three clusters, those in the range from $10^{-8}$ to $10^{-6}$, namely BatchNorm, InstanceNorm and LayerNorm-AF, one with a larger error in the range from $10^{-4}$ to $10^{-3}$, namely LayerNorm-C, and one with the largest error in the range from $10^{-2}$ to $10^{-1}$, namely LayerNorm-CHW. The results suggest that BatchNorm and LayerNorm-AF are equivariant to shifts and translations, that InstanceNorm and LayerNorm-C are equivariant to shifts but not to translations, and that LayerNorm-CHW is equivariant to neither.

The results are consistent with our theoretical predictions as LayerNorm-CHW is the only one with a large shift-equivariance error, as BatchNorm, InstanceNorm and LayerNorm-AF have a low translation-equivariance error, and as LayerNorm-CHW and LayerNorm-C have a high translation-equivariance error. We believe that the difference between LayerNorm-C and LayerNorm-CHW can be understood as showing that the affine step hinders equivariance significantly more than the scaling step.

Batch normalization operates in two different modes: training mode, and evaluation mode. \Cref{tbl:batch_norm} shows the equivariance error to shifts and translations for both modes. For shift-equivariance and translation-equivariance in evaluation mode, the equivariance error is in the order of $10^{-9}$, and for translation-equivariance in training mode, it is in the order of $10^{-6}$. All four values are fairly low, which is coherent with our theoretical predictions, but is is not entirely clear why one of the values is higher than the others. In comparison to the results in \Cref{tbl:main}, the lowest value is more consistent with the other normalization layers, which might indicate that most of the equivariance error is due to the non-linear normalization, as opposed to the linear normalization done using running statistics.

Additionally, we confirm the role of aliasing in the equivariance to translations in~\Cref{section.additional_experiments}.

\begin{table}[htbp]
	\centering
	\setlength{\tabcolsep}{28pt} % default: 6pt
	\begin{tabular}{lll}
	\toprule
	 & Shifts & Translations \\
	Normalization mode &  &  \\
	\midrule
	Training & \textbf{9.51e-09 ± 3.79e-12} & \textbf{2.55e-06 ± 3.15e-09} \\
	Evaluation & \textbf{8.95e-09 ± 4.26e-12} & \textbf{9.69e-09 ± 3.46e-12} \\
	\bottomrule
	\end{tabular}
	\caption{\textbf{Equivariance error for the two modes of batch normalization.}
		Batch normalization behaves differently in training and evaluation modes. In the experiments, we randomize the mode and measure the equivariance error for each mode. The results are consistent with the theoretical predictions as shown by low equivariance error for shifts and translations. In \textbf{bold}, values lower than $10^{-4}$. Values: avg $\pm$ s.e.}
	\label{tbl:batch_norm}
	\vspace{-1cm}
\end{table}

\section{Conclusion}
\label{section.conclusion}

In this work, we study the equivariance of normalization layers to shifts and translations. Our new theoretical framework highlights that the dimensions the affine step operates on, and the way the standard deviation is computed are the two factors determining whether a normalization layer is equivariant or not to shifts and/or translations. More precisely, we prove that shift-equivariant normalization layers are those with an affine step that does not operate on the spatial dimensions, or with no affine step altogether, and that translation-equivariant further requires that the standard deviation be computed at least on the spatial dimensions. We test our theoretical predicitions empirically by measuring and comparing the equivariance error to shifts and translations of five common normalization layers, and obtain results that are consistent with the predictions.

The choice of a normalization layer affects not only the equivariance of the overall neural architecture, but also its performance first and foremost. In this work, we focus on the equivariance properties of normalization layers, and while equivariance and performance tend to be correlated in well-designed architectures, we emphasize that empirical validation of the performance is crucial to the selection of a normalization layer. Studying the relation between equivariance and performance in normalization layers is an exciting research direction, and we leave it for future work.

Vision transformers are an important family of neural architectures used for computer vision. Yet, their equivariance to continuous translations remains to be studied and more groundwork is needed before the equivariance of their normalization layers can be studied specifically. Until then, it is unclear whether our theoretical results will generalize to them, or if they will remain valid only for convolutional architectures. We believe that this would also constitute an interesting research question for future work.

\bibliography{/home/j.scanvic@fox.cos/Documents/work/refs/All}

\begin{thebibliography}{24}
\providecommand{\natexlab}[1]{#1}
\providecommand{\url}[1]{\texttt{#1}}
\expandafter\ifx\csname urlstyle\endcsname\relax
  \providecommand{\doi}[1]{doi: #1}\else
  \providecommand{\doi}{doi: \begingroup \urlstyle{rm}\Url}\fi

\bibitem[Azulay and Weiss(2019)]{azulay19Why}
Aharon Azulay and Yair Weiss.
\newblock Why do deep convolutional networks generalize so poorly to small image transformations?, December 2019.

\bibitem[Ba et~al.(2016)Ba, Kiros, and Hinton]{ba16Layer}
Jimmy~Lei Ba, Jamie~Ryan Kiros, and Geoffrey~E. Hinton.
\newblock Layer {{Normalization}}, July 2016.

\bibitem[Chaman and Dokmani{\'c}(2021{\natexlab{a}})]{chaman21Truly}
Anadi Chaman and Ivan Dokmani{\'c}.
\newblock Truly shift-invariant convolutional neural networks, March 2021{\natexlab{a}}.

\bibitem[Chaman and Dokmani{\'c}(2021{\natexlab{b}})]{chaman21TrulyEquivariant}
Anadi Chaman and Ivan Dokmani{\'c}.
\newblock Truly shift-equivariant convolutional neural networks with adaptive polyphase upsampling, December 2021{\natexlab{b}}.

\bibitem[Cohen and Welling(2016{\natexlab{a}})]{cohen16Group}
Taco~S. Cohen and Max Welling.
\newblock Group {{Equivariant Convolutional Networks}}, June 2016{\natexlab{a}}.

\bibitem[Cohen and Welling(2016{\natexlab{b}})]{cohen16Steerable}
Taco~S. Cohen and Max Welling.
\newblock Steerable {{CNNs}}, December 2016{\natexlab{b}}.

\bibitem[Deng et~al.(2009)Deng, Dong, Socher, Li, Li, and {Fei-Fei}]{deng09ImageNet}
Jia Deng, Wei Dong, Richard Socher, Li-Jia Li, Kai Li, and Li~{Fei-Fei}.
\newblock {{ImageNet}}: {{A}} large-scale hierarchical image database.
\newblock In \emph{2009 {{IEEE Conference}} on {{Computer Vision}} and {{Pattern Recognition}}}, pages 248--255, June 2009.
\newblock \doi{10.1109/CVPR.2009.5206848}.

\bibitem[Gruver et~al.(2024)Gruver, Finzi, Goldblum, and Wilson]{gruver24Lie}
Nate Gruver, Marc Finzi, Micah Goldblum, and Andrew~Gordon Wilson.
\newblock The {{Lie Derivative}} for {{Measuring Learned Equivariance}}, June 2024.

\bibitem[He et~al.(2015)He, Zhang, Ren, and Sun]{he15Deep}
Kaiming He, Xiangyu Zhang, Shaoqing Ren, and Jian Sun.
\newblock Deep {{Residual Learning}} for {{Image Recognition}}, December 2015.

\bibitem[Ioffe and Szegedy(2015)]{ioffe15Batch}
Sergey Ioffe and Christian Szegedy.
\newblock Batch {{Normalization}}: {{Accelerating Deep Network Training}} by {{Reducing Internal Covariate Shift}}, March 2015.

\bibitem[Karras et~al.(2021)Karras, Aittala, Laine, H{\"a}rk{\"o}nen, Hellsten, Lehtinen, and Aila]{karras21AliasFree}
Tero Karras, Miika Aittala, Samuli Laine, Erik H{\"a}rk{\"o}nen, Janne Hellsten, Jaakko Lehtinen, and Timo Aila.
\newblock Alias-{{Free Generative Adversarial Networks}}, October 2021.

\bibitem[Kim et~al.(2023)Kim, Baucour, and Shin]{kim23ModelAgnostic}
Myungjoon Kim, Arthur Baucour, and Jonghwa Shin.
\newblock Model-{{Agnostic Shift-Equivariant Downsampling}}.
\newblock October 2023.

\bibitem[Liu et~al.(2022)Liu, Mao, Wu, Feichtenhofer, Darrell, and Xie]{liu22ConvNet}
Zhuang Liu, Hanzi Mao, Chao-Yuan Wu, Christoph Feichtenhofer, Trevor Darrell, and Saining Xie.
\newblock A {{ConvNet}} for the 2020s, March 2022.

\bibitem[Michaeli et~al.(2023)Michaeli, Michaeli, and Soudry]{michaeli23AliasFree}
Hagay Michaeli, Tomer Michaeli, and Daniel Soudry.
\newblock Alias-{{Free Convnets}}: {{Fractional Shift Invariance}} via {{Polynomial Activations}}, March 2023.

\bibitem[Ravanbakhsh et~al.(2017)Ravanbakhsh, Schneider, and Poczos]{ravanbakhsh17Equivariance}
Siamak Ravanbakhsh, Jeff Schneider, and Barnabas Poczos.
\newblock Equivariance {{Through Parameter-Sharing}}, June 2017.

\bibitem[{Rojas-Gomez} et~al.(2022){Rojas-Gomez}, Lim, Schwing, Do, and Yeh]{rojas-gomez22Learnable}
Renan~A. {Rojas-Gomez}, Teck-Yian Lim, Alexander~G. Schwing, Minh~N. Do, and Raymond~A. Yeh.
\newblock Learnable {{Polyphase Sampling}} for {{Shift Invariant}} and {{Equivariant Convolutional Networks}}, October 2022.

\bibitem[Ruzanski and Chandrasekar(2011)]{ruzanski11Scale}
Evan Ruzanski and V.~Chandrasekar.
\newblock Scale {{Filtering}} for {{Improved Nowcasting Performance}} in a {{High-Resolution X-Band Radar Network}}.
\newblock \emph{IEEE Transactions on Geoscience and Remote Sensing}, 49\penalty0 (6):\penalty0 2296--2307, June 2011.
\newblock ISSN 1558-0644.
\newblock \doi{10.1109/TGRS.2010.2103946}.

\bibitem[Saha and Gokhale(2024)]{saha24Improving}
Sourajit Saha and Tejas Gokhale.
\newblock Improving {{Shift Invariance}} in {{Convolutional Neural Networks}} with {{Translation Invariant Polyphase Sampling}}, December 2024.

\bibitem[Ulyanov et~al.(2017)Ulyanov, Vedaldi, and Lempitsky]{ulyanov17Instance}
Dmitry Ulyanov, Andrea Vedaldi, and Victor Lempitsky.
\newblock Instance {{Normalization}}: {{The Missing Ingredient}} for {{Fast Stylization}}, November 2017.

\bibitem[Vetterli et~al.(2014)Vetterli, Kova{\v c}evi{\'c}, and Goyal]{vetterli14Foundations}
Martin Vetterli, Jelena Kova{\v c}evi{\'c}, and Vivek~K Goyal.
\newblock \emph{Foundations of Signal Processing}.
\newblock Cambridge University Press, 2014.

\bibitem[Weiler and Cesa(2021)]{weiler21General}
Maurice Weiler and Gabriele Cesa.
\newblock General \${{E}}(2)\$-{{Equivariant Steerable CNNs}}, April 2021.

\bibitem[Wu and He(2018)]{wu18Group}
Yuxin Wu and Kaiming He.
\newblock Group {{Normalization}}, June 2018.

\bibitem[Zhang(2019)]{zhang19Making}
Richard Zhang.
\newblock Making {{Convolutional Networks Shift-Invariant Again}}, June 2019.

\bibitem[Zou et~al.(2020)Zou, Xiao, Yu, and Lee]{zou20Delving}
Xueyan Zou, Fanyi Xiao, Zhiding Yu, and Yong~Jae Lee.
\newblock Delving {{Deeper}} into {{Anti-aliasing}} in {{ConvNets}}, August 2020.

\end{thebibliography}

% Appendix %
\appendix

\begin{figure}[htbp]
	\centering
	\includegraphics[width=\textwidth]{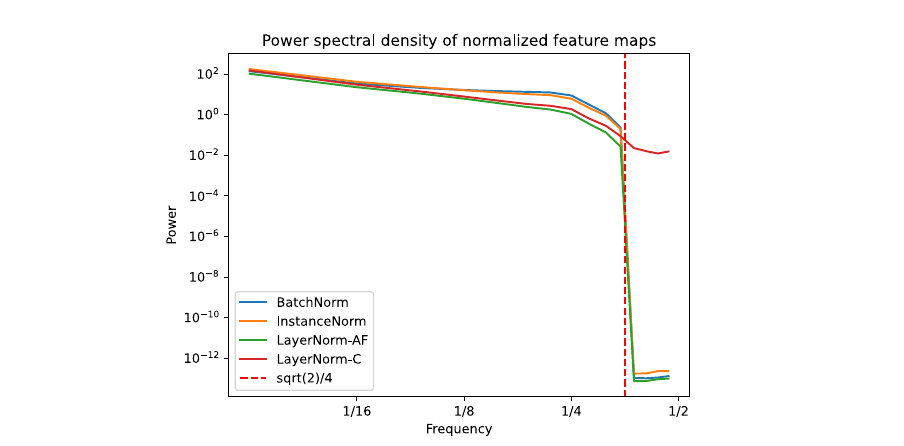}
	\caption{\textbf{Detection of aliasing in normalization layers.} In~\Cref{section:analysis}, we explain that the layers that are equivariant to discrete shifts are also equivariant to translations if, and only if, they are not prone to aliasing. Here, we show the radial power spectral density of the normalized feature maps obtained from $\times 2$ upsampled inputs with no energy in the band $(\frac{\sqrt 2}{4}, \frac 1 2)$. Energy in this band indicates aliasing due to an increased frequency bandwidth.}
	\label{fig:aliasing}
\end{figure}

\section{Additional experiments}
\label{section.additional_experiments}

In~\Cref{section:analysis}, we show theoretically that certain normalization
layers that are equivariant to discrete shifts are prone to aliasing and thus
not equivariant to continuous translations, and in~\Cref{section.experiments},
we show empirically that the normalization layers that are predicted to be
equivariant to translations have a low equivariance error to translations, and
that those that are predicted to not be equivariant to translations have a high
equivariance error to translations. In this section, we show empirically that
aliasing is indeed what distinguishes layers that are simply equivariant to
shifts from those that are also equivariant to translations.

Aliasing is the spectral folding of high-frequency information into the lower
frequency range caused, in the case of normalization layers, by an increase of
the actual signal bandwidth without a proper increase in sampling rate. We
propose to detect it using the tools from spectral analysis. More precisely, we
consider the same feature maps we used in~\Cref{section.experiments} and we
upsample them by a factor of $2$ using an ideal sinc low-pass filter to get rid
of the higher frequencies while still leaving room for them. Then, we apply
each of the 4 normalization layers that are equivariant to shifts, namely
BatchNorm, InstanceNorm, LayerNorm-C and LayerNorm-AF. Presence of energy above
the cut-off frequency of the sinc filter indicates that the layer increases the
effective bandwidth of its input, and thus that it is prone to aliasing.

We compute the radial power spectral density (PSD)~\citep{ruzanski11Scale} over
all of the normalized feature maps for each layer and see if there is energy in
the aliasing frequency band ranging from $\frac{\sqrt 2}{4}$ to $\frac 1 2$.
\Cref{fig:aliasing} shows that most of the layers have barely any energy in the
aliasing band, with a power spectral density of about $10^{-12}$ in that range,
and that the remaining one does have some energy, with a power spectral density
of about $10^{-2}$. Moreover, the layers without energy in the aliasing band
are exactly those predicted to be equivariant to translations and with a low
empirical equivariance error to translations. Overall, the results are
consistent with our theoretical and empirical results.

\section{Proofs}
\label{section:proofs}

We make a few simplifications to make the proof easier to follow. Instead of considering 2-dimensional feature maps, we consider 1-dimension feature maps, and instead of considering feature maps with separate batch and channel dimensions, we consider a single batch/channel dimension. We believe that the proof lays most of the groundwork to prove the general case as the two spatial dimensions can be treated similarly, and the batch and channel dimensions as well.

Formally, normalization layers are functions $f_{\gamma,\beta} : \mathbb R^{K \times D} \to \mathbb R^{K \times D}$, where $K \geq 1$ is the number of spatial dimensions and $D \geq 1$ is the number of batch and channel dimensions, and where $(\gamma, \beta) \in \Theta \subseteq \mathbb R^{K \times D} \times \mathbb R^{K \times D}$ represent the learned affine transform parameters. The set $\Theta$ represents the set of admissible affine transforms corresponding to a given normalization layer. For instance, normalization layers without an affine transform are modelled as
\begin{equation}
	\Theta_0 = \{(\mathbf{1}_{K \times D}, \mathbf{0}_{K \times D}) \}
\end{equation}
 where $\mathbf{1}_{K \times D}$ is the $K \times D$ matrix of ones and $\mathbf{0}_{K \times D}$ is the $K \times D$ matrix of zeros, which forces the affine transform to be the identity. Affine transforms restrained to batch/channel dimensions are modelled as
\begin{equation}
	\Theta_D = \{ (\gamma, \beta) \in \mathbb R^{K \times D} \times \mathbb R^{K \times D}, \gamma_{kd} = \gamma_{k'd}, \beta_{kd} = \beta_{k'd}, \forall k, k', d \}
\end{equation}
and those restricted to spatial dimensions are modelled as
\begin{equation}
	\Theta_K = \{ (\gamma, \beta) \in \mathbb R^{K \times D} \times \mathbb R^{K \times D}, \gamma_{kd} = \gamma_{kd'}, \beta_{kd} = \beta_{kd'}, \forall k, d, d' \},
\end{equation}
that is, as matrices with equal rows or columns.

Instead of considering all eight cases for centering and scaling on no/one/two dimensions each, we focus on the two most important cases: centering and scaling on the batch/channel dimension, or on the spatial dimension. Taking into account those simplifications, normalization layers are expressed as
\begin{equation}
	f_{\gamma,\beta}(x) = \gamma \odot \frac{x - \mathbb E[x]}{\sqrt{\mathbb E[|x - \mathbb E[x]|^2]}} + \beta, \ x \in \mathbb R^{K \times D},
\end{equation}
where the expectation is computed over the batch/channel dimension
\begin{equation}
	\mathbb E_d[x_{kd}] := \frac 1 D \sum_{d'=0}^{D-1} x_{kd'}, \ x \in \mathbb R^{K \times D},
\end{equation}
or over the spatial dimension
\begin{equation}
	\mathbb E_k[x_{kd}] := \frac 1 K \sum_{k'=0}^{K-1} x_{k'd}, \ x \in \mathbb R^{K \times D}.
\end{equation}

In this setting, the translation operation is one-dimensional and is defined as
\begin{equation}
	T_g x_{kd} = \mathrm{IDFT}_1\left(e^{-i 2 \pi \frac{kg}{K}} (\mathrm{DFT}_1(x_{kd}))\right),\ x \in \mathbb R^{K \times D},
\end{equation}
where $g \in \mathbb R$ is the displacement, and $\mathrm{DFT}_1$ and $\mathrm{IDFT}_1$ are the 1-dimensional discrete Fourier transform and its inverse, respectively. As a diagonal operator in the Fourier basis, it is also a convolutional operator
\begin{equation}
	T_g x_{kd} = \varphi_{g,k} * x_{k,d}= \sum_{k'=0}^{K-1} x_{dk'} \varphi_{g,(k-k')_K}
\end{equation}
where $*$ denotes spatial 1-dimensional convolution, and whose kernel is expressed as
\begin{equation} \label{eq:translation_kernel}
	\varphi_{g,k} = \begin{cases}
		\frac 1 K \frac{\sin(\pi (g - k))}{\sin(\pi (g - k) / K)} e^{-i\pi (g - k) \left( 1 - \frac 1 K \right)}, &\text{if $g \in \mathbb R \setminus \mathbb Z$} \\
		\delta_{kg} &\text{if $g \in \mathbb Z$},
\end{cases}
\end{equation}
where $\delta_{kg}$ is the Kronecker symbol.

\subsection{Preliminaries}

The averaging operation is linear in all cases, and $T_g$ is a linear operator for all $g \in \mathbb R$, so
\begin{equation}
	T_g x - \mathbb E[T_g x] = T_g x - T_g \mathbb E[x] = T_g (x - \mathbb E[x]),\ g \in \mathbb R.
\end{equation}
That is, centering never causes a lack of equivariance to translations (or shifts). This property is particularly important and we use it extensively through the rest of the proof.

\subsection{Proof of Theorem~\ref{theorem:shifts}}

For convenience, we restate the theorem:

\theoremShifts*

We prove the theorem in two steps: i) we show that if the affine step operates on the spatial dimension, then the normalization layer is not equivariant to shifts, and ii) we show that if there is no affine step or if the affine step does not operate on the spatial dimension, then the normalization layer is equivariant to shifts.

\paragraph{Non-equivariance to shifts}

Let's assume that the affine step operates on the spatial dimensions, i.e., that $\Theta_K \subseteq \Theta$. Let $\gamma_{kd} = \delta_{k0}$, and $\beta = \mathbf{0}_{K \times D}$, since $\gamma$ and $\beta$ do not vary along the batch/channel dimension $D$, $(\gamma, \beta) \in \Theta_D \subseteq \Theta$ and it is an admissible set of parameters for the normalization layer.

Let $g \in \mathbb Z \setminus \{ 0 \}$ and $x_{kd} \in \mathbb R^{K \times D}$ be any feature map with normalization
\begin{equation}
	y_{kd} = \frac{x_{kd} - \mathbb E[x_{kd}]}{\sqrt{\mathbb E[|x_{kd} - \mathbb E[x_{kd}]|^2]}}
\end{equation}
having no entry equal to zero
\begin{equation}
	y_{kd} \neq 0, \forall k = 0, \ldots, K - 1, \forall d = 0, \ldots, D - 1.
\end{equation}
Note that this can be achieved by letting $x_{kd}$ take its values in $\{ -1, 1 \}$ with at least one occurrence of the two numbers on each row. Indeed, in that case, the mean takes values in $(-1, 1)$, and the centered feature map has no entry equal to zero. And since scaling cannot introduce any new zero, the normalized feature map has no entry equal to zero either.

\begin{align}
	f_{\gamma,\beta}(T_g x_{kd}) - T_g f_{\gamma,\beta}(x_{kd}) &= \gamma_{kd} \odot \frac{T_g x_{kd} - \mathbb E[T_g x_{kd}]}{\sqrt{\mathbb E[|T_g x_{kd} - \mathbb E[T_g x_{kd}]|^2]}} + \beta_{kd} \\
								    &- T_g \left( \gamma_{kd} \odot \frac{x_{kd} - \mathbb E[x_{kd}]}{\sqrt{\mathbb E[|x_{kd} - \mathbb E[x_{kd}]|^2]}} + \beta_{kd}\right)\\
								    &= \gamma_{kd} \odot T_g\left( \frac{x_{kd} - \mathbb E[x_{kd}]}{\sqrt{\mathbb E[|x_{kd} - \mathbb E[x_{kd}]|^2]}} \right) \\
								    &- T_g \left( \gamma_{kd} \odot \frac{x_{kd} - \mathbb E[x_{kd}]}{\sqrt{\mathbb E[|x_{kd} - \mathbb E[x_{kd}]|^2]}} \right)\\
								    &= \gamma_{kd} \odot T_g\left( \frac{x_{kd} - \mathbb E[x_{kd}]}{\sqrt{\mathbb E[|x_{kd} - \mathbb E[x_{kd}]|^2]}} \right) \\
								    &- (T_g \gamma_{kd}) \odot \underbrace{T_g \left(\frac{x_{kd} - \mathbb E[x_{kd}]}{\sqrt{\mathbb E[|x_{kd} - \mathbb E[x_{kd}]|^2]}} \right)}_{y_{kd}} \\
								    &= \delta_{k0} y_{kd} - \delta_{kg} y_{kd} \\
								    &= \begin{cases}
									    y_{0d} &\text{if $k = 0$},\\
									    -y_{gd} &\text{if $k = g$}\\
									    0 &\text{otherwise.}
								    \end{cases}
\end{align}

Since $y_{kd}$ has no entry equal to zero, the equivariance error has non-zero entries and thus the normalization layer is not equivariant to shifts.

\paragraph{Equivariance to shifts}

Let's assume that there is no affine step, or that it does not operate on the spatial dimensions, i.e., that $\Theta \subseteq \Theta_D$. Let $(\gamma, \beta) \in \Theta$, $g \in \mathbb Z$, and $x_{kd} \in \mathbb R^{K \times D}$. Since $(\gamma, \beta) \in \Theta_D$, we have
\begin{equation}
	T_g \gamma_{kd} = \gamma_{kd},\ T_g \beta_{kd} = \beta_{kd}.
\end{equation}

\begin{align}
	f_{\gamma,\beta}(T_g x_{kd}) &= \gamma_{kd} \odot \frac{T_g x_{kd} - \mathbb E[T_g x_{kd}]}{\sqrt{\mathbb E[|T_g x_{kd} - \mathbb E[T_g x_{kd}]|^2]}} + \beta_{kd} \\
				     &= \gamma_{kd} \odot T_g \left( \frac{x_{kd} - \mathbb E[x_{kd}]}{\sqrt{\mathbb E[|x_{kd} - \mathbb E[x_{kd}]|^2]}} \right) + \beta_{kd} \\
				     &= (T_g \gamma_{kd}) \odot T_g \left( \frac{x_{kd} - \mathbb E[x_{kd}]}{\sqrt{\mathbb E[|x_{kd} - \mathbb E[x_{kd}]|^2]}} \right) + (T_g \beta_{kd}) \\
				     &= T_g \left(\gamma_{kd} \odot  \frac{x_{kd} - \mathbb E[x_{kd}]}{\sqrt{\mathbb E[|x_{kd} - \mathbb E[x_{kd}]|^2]}} + \beta_{kd} \right) \\
				     &= T_g f_{\gamma,\beta}(x_{kd}).
\end{align}

Since this is true for all $\gamma, \beta, g$, and $x$, the normalization layer is equivariant to shifts.

\subsection{Proof of Theorem~\ref{theorem:translations}}

Again, we restate the theorem:

\theoremTranslations*

Recall that shift-equivariance is the same as translation-equivariance except restrained to whole pixel displacements. This makes shift-equivariance a necessary condition for translation-equivariance. It suffices to show that a shift-equivariant layer is also translation-equivariant if, and only if, the standard deviation is computed on the spatial dimensions.

We prove that in two steps: i) we show that if the standard deviation is not computed on spatial dimensions, then the normalization layer is not equivariant to translations, and ii) we show that if the standard deviation is computed on the spatial dimensions, then the normalization layer is equivariant to translations.

In this section, we assume that the normalization layer is equivariant to shifts, or equivalently, according to~\Cref{theorem:shifts}, that there is no affine step or that it operates on the batch/channel dimensions $\Theta \subseteq \Theta_D$.

\paragraph{Non-equivariance to translations}

Let's assume that the scaling is done over the batch/channels dimension
\begin{equation}
	\mathbb E[x_{kd}] = \mathbb E_d[x_{kd}] := \frac 1 D \sum_{d'=0}^{D-1} x_{kd'}.
\end{equation}
We let $x_{kd} = \delta_{0k} u_{d}$ where $\delta$ is the Kronecker delta and $u_d \in \mathbb R^D$ is any vector with mean zero and variance one. We also let $g \in \mathbb R \setminus \mathbb Z$ and $\gamma = \mathbf{1}_{K \times D}$ and $\beta = \mathbf{0}_{K \times D}$. We compute:

\begin{align}
	f_{\gamma,\beta}(T_g x_{kd}) - T_g f_{\gamma,\beta}(x_{kd}) &= \gamma_{kd} \odot \frac{T_g x_{kd} - \mathbb E[T_g x_{kd}]}{\sqrt{\mathbb E[|T_g x_{kd} - \mathbb E[T_g x_{kd}]|^2]}} + \beta_{kd} \\
								    & - T_g \left( \gamma_{kd} \odot \frac{x_{kd} - \mathbb E[x_{kd}]}{\sqrt{\mathbb E[|x_{kd} - \mathbb E[x_{kd}]|^2]}} + \beta_{kd}\right)\\
								    &= \frac{T_g x_{kd} - T_g \mathbb E[x_{kd}]}{\sqrt{\mathbb E[|T_g x_{kd} - T_g \mathbb E[ x_{kd}]|^2]}} - T_g \left( \frac{x_{kd}}{\sqrt{\mathbb E[|x_{kd}|^2]}} \right)\\
								    &= \frac{T_g x_{kd}}{\sqrt{\mathbb E[|T_g x_{kd}|^2]}} - T_g \left( \frac{x_{kd}}{\sqrt{\mathbb E[|x_{kd}|^2]}} \right)\\
	&= (\varphi_{gk} * x_{kd}) \odot \mathbb E_d[(\varphi_{gk} * x_{kd})^2]^{-1/2} \\
	&- \varphi_{gk} * \left( x_{kd} \odot \mathbb E_d[x_{kd}^2]^{-1/2} \right) \\
	&= (\varphi_{gk} * \delta_{0k} u_d) \odot \mathbb E_d[(\varphi_{gk} * \delta_{0k} u_{d})^2]^{-1/2} \\
	&- \varphi_{gk} * \left( \delta_{0k} u_{d} \odot \mathbb E_d[\delta_{0k} u_d^2]^{-1/2} \right) \\
	&= \varphi_{gk} u_d \odot |\varphi_{gk}|^{-1} \mathbb E_d[u_{d}^2]^{-1/2} \\
	&- \varphi_{gk} * \left( \delta_{0k} u_{d} \odot \delta_{0k} \mathbb E_d[u_{d}^2]^{-1/2} \right) \\
	&= \varphi_{gk} u_d \odot |\varphi_{gk}|^{-1} - \varphi_{gk} * \left( \delta_{0k} u_{d} \odot \delta_{0k} \right) \\
	&= \varphi_{gk} u_d \odot |\varphi_{gk}|^{-1} - \varphi_{gk} * u_d \\
	&= \varphi_{gk} \left( \frac{1}{|\varphi_{gk}|} - 1 \right) u_d. \\
\end{align}
By definition, $u_d \neq 0$ and, according to~\cref{eq:translation_kernel}, $|\varphi_{gk}|$ is not in $\{ 0, 1 \}$ for all $k$, as $g \in \mathbb R \setminus \mathbb Z$, so the equivariance error is non-zero. The normalization layer is not equivariant to translations.

\paragraph{Equivariance to translations}

Let's assume that the scaling is done over the spatial dimensions
\begin{equation}
	\mathbb E[x_{kd}] = \mathbb E_k[x_{kd}] := \frac 1 K \sum_{k'=0}^{K-1} x_{k'd}.
\end{equation}
Let $(\gamma, \beta) \in \Theta$, $x \in \mathbb R^{K \times D}$ be any signal, $g \in \mathbb R$. Since $\Theta \subseteq \Theta_D$, we have
\begin{equation}
	T_g \gamma_{kd} = \gamma_{kd},\ T_g \beta_{kd} = \beta_{kd}.
\end{equation}

We first show that $T_g$ is unitary
\begin{equation}
	\mathbb E_k[(T_g x_{kd})^2] = \mathbb E[x_{kd}^2].
\end{equation}
Indeed, since the DFT is unitary~\citep{vetterli14Foundations}, we have
\begin{align}
	\mathbb E_k[(T_g x_{kd})^2] &= \frac 1 K \sum_{k=0}^{K-1} (T_g x_{kd})^2 \\
				    &= \frac 1 K \sum_{k=0}^{K-1} \mathrm{DFT}_1(T_g x_{kd})^2 \\
				    &= \frac 1 K \sum_{k=0}^{K-1} |e^{-i 2 \pi \frac{kg}{K}} \mathrm{DFT}_1(x_{kd})|^2 \\
				    &= \frac 1 K \sum_{k=0}^{K-1} |\mathrm{DFT}_1(x_{kd})|^2 \\
				    &= \frac 1 K \sum_{k=0}^{K-1} x_{kd}^2 \\
				    &= \mathbb E[x_{kd}^2].
\end{align}

Now, we compute
\begin{align}
	f_{\gamma,\beta}(T_g x_{kd}) - T_g f_{\gamma,\beta}(x_{kd}) &= \gamma_{kd} \odot \frac{T_g x_{kd} - \mathbb E[T_g x_{kd}]}{\sqrt{\mathbb E[|T_g x_{kd} - \mathbb E[T_g x_{kd}]|^2]}} + \beta_{kd} \\
								    & - T_g \left( \gamma_{kd} \odot \frac{x_{kd} - \mathbb E[x_{kd}]}{\sqrt{\mathbb E[|x_{kd} - \mathbb E[x_{kd}]|^2]}} + \beta_{kd}\right)\\
								    &= \gamma_{kd} \odot \frac{T_g x_{kd} - T_g \mathbb E[x_{kd}]}{\sqrt{\mathbb E[|T_g x_{kd} - T_g \mathbb E[ x_{kd}]|^2]}} \\
								    &- T_g \gamma_{kd} \odot T_g \left( \frac{x_{kd} - \mathbb E[x_{kd}]}{\sqrt{\mathbb E[|x_{kd} - \mathbb E[x_{kd}]|^2]}} \right) + \beta_{kd} - T_g \beta_{kd} \\
								    &=\gamma_{kd} \odot \frac{T_g (x_{kd} - \mathbb E[x_{kd}])}{\sqrt{\mathbb E[|T_g (x_{kd} - \mathbb E[ x_{kd}])|^2]}} \\
								    &- \gamma_{kd} \odot T_g \left( \frac{x_{kd} - \mathbb E[x_{kd}]}{\sqrt{\mathbb E[|x_{kd} - \mathbb E[x_{kd}]|^2]}} \right) \\
								    &=\gamma_{kd} \odot \Bigg( \frac{T_g (x_{kd} - \mathbb E[x_{kd}])}{\sqrt{\mathbb E[|x_{kd} - \mathbb E[ x_{kd}]|^2]}} - T_g \left( \frac{x_{kd} - \mathbb E[x_{kd}]}{\sqrt{\mathbb E[|x_{kd}|^2]}} \right) \Bigg) \\
								    &=\gamma_{kd} \odot \Bigg( \Big( \varphi_{gk} *_k (x_{kd} - \mathbb E[x_{kd}]) \Big) \odot_d \mathbb E_k[|x_{kd} - \mathbb E[ x_{kd}]|^2]^{-1/2} \\
								    &- \varphi_{gk} * \Big( (x_{kd} - \mathbb E[x_{kd}]) \odot_d \mathbb E_k[|x_{kd} - \mathbb E[x_{kd}]|^2]^{-1/2} \Big) \Bigg) \\
								    &= 0
\end{align}
where $*_k$ emphasizes convolution on the $k$ dimension, and $\odot_d$ emphasizes Hadamard product on the $d$ dimension. The equivariance error is zero, and thus the normalization layer is equivariant to translations.

\end{document}